\documentclass[letterpaper, 10 pt, conference]{ieeeconf}
\IEEEoverridecommandlockouts
\usepackage{epsfig}
\usepackage{amssymb}
\usepackage{cite}
\usepackage{booktabs}
\usepackage{multirow}
\usepackage{mathtools}
\usepackage{bm}
\usepackage{algorithm}
\usepackage{comment}
\usepackage{color}
\usepackage{setspace}

\newcommand{\figlab}[1]{\label{fig:#1}}
\newcommand{\figref}[1]{Fig.~\ref{fig:#1}} 
\newcommand{\tablab}[1]{\label{tab:#1}}
\newcommand{\tabref}[1]{Table~\ref{tab:#1}} 
\newcommand{\forlab}[1]{\label{for:#1}}
\newcommand{\forref}[1]{Equation~(\ref{for:#1})} 

\newcommand{\etal}{\textit{et al.~}}

\definecolor{green}{rgb}{0.01, 0.5, 0.01}

\usepackage{subcaption}
\usepackage{url}
\usepackage{threeparttable}

\usepackage{amsmath}
\usepackage[noend]{algorithmic}
\usepackage{array}

\begin{document}
\title{Task-Difficulty-Aware Efficient Object Arrangement \\Leveraging Tossing Motions}
\author{Takuya Kiyokawa$^{*}$, Mahiro Muta, Weiwei Wan, and Kensuke Harada
\thanks{This research is subsidized by the New Energy and Industrial Technology Development Organization (NEDO) under project JPNP20016. This paper is one of the achievements of joint research with and is jointly owned copyrighted material of ROBOT Industrial Basic Technology Collaborative Innovation Partnership.}
\thanks{All authors are with the Department of Systems Innovation, Graduate School of Engineering Science, Osaka University, Toyonaka, Osaka, Japan. $^{*}$Corresponding author: {\tt\small kiyokawa@sys.es.osaka-u.ac.jp}}%
}

\maketitle

\vspace{-20pt}
\begin{abstract}
This study explores a pick-and-toss (PT) as an alternative to pick-and-place (PP), allowing a robot to extend its range and improve task efficiency. Although PT boosts efficiency in object arrangement, the placement environment critically affects the success of tossing. To achieve accurate and efficient object arrangement, we suggest choosing between PP and PT based on task difficulty estimated from the placement environment. Our method simultaneously learns the tossing motion through self-supervised learning and the task determination policy via brute-force search. Experimental results validate the proposed method through simulations and real-world tests on various rectangular object arrangements.
\end{abstract}

\IEEEpeerreviewmaketitle

\section{Introduction}
Pick-and-toss (PT) can be more effective than pick-and-place (PP) for robotic object arrangement operations in the retail industry. Although several methods have been proposed to achieve accurate dynamic object manipualtion motions like tossing motions~\cite{Senoo2008,Miyashita2010,Kajikawa2014,Takahashi2021}, object placement in an arbitrary posture is not considered. Achieving both accuracy and efficiency in object arrangement using PT alone is difficult due to the placement environment's significant effect on the success of the tossing motion.

This study aims to achieve accurate and efficient object arrangement by addressing the task determination problem of PP and PT, considering task difficulty based on the number and type of contact surfaces in the target placement environment. We propose a framework where the robot learns the tossing motion and task determination policy through trial and error, introducing a method to acquire the tossing motion using self-supervised learning and the task determination policy through brute-force search. Our experiments validate the method's effectiveness in both simulations and real-world scenarios.
\begin{figure}[tb]
    \centering
    \includegraphics[keepaspectratio, width=\linewidth]{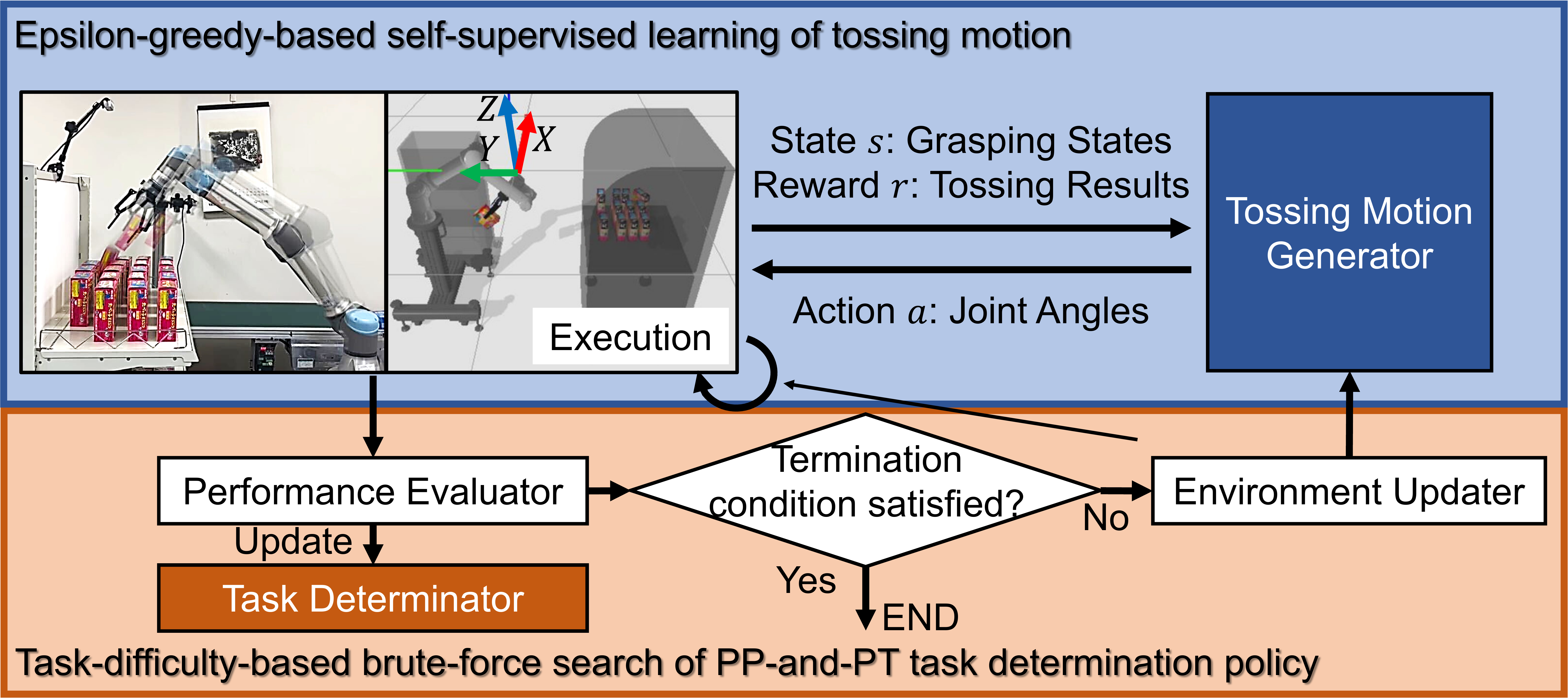}
    \caption{\small{Training pipeline for PP-and-PT-based object arrangement.}}
    \figlab{overview}
\end{figure} 

\section{Proposed Method}
\subsection{Problem Setting and Overview}
In this study, the PP and PT tasks are performed using a robotic arm with six joints, equipped with a two-finger gripper, focusing on rectangular shapes. An RGB-D sensor above the workspace captures images of the objects, and their regions and category names are identified using the instance segmentation method, Detic~\cite{Zhou2022}. A bounding box is created from the object's contour, and its center is calculated. Based on the center position, locations without objects are identified, and a suitable placement location for the grasped object is chosen.

We propose a method to simultaneously acquire the PP and PT task determination policy (Task Determinator) and generate tossing motions based on the grasping state (Tossing Motion Generator), as shown in~\figref{overview}.

\subsection{Task Determination Based on Contact Surfaces}
\subsubsection{Defining Contact Patterns}
Yoshikawa~\etal~\cite{Yoshikawa1991} introduced the concept of constraint state transition difficulty, stating that an increase in constraint degree, determined by the contact state, corresponds to a higher difficulty. This study assumes that task difficulty varies based on contact pattern (C-pattern) defined by the number and type of contact surfaces—specifically, the number of contact surfaces with other objects and whether these surfaces are movable or fixed.

The C-pattern illustrated in~\figref{cp} is determined by the number and type of contacts between the target object and floor, walls, and preplaced objects. In the figure, orange represents the target object, blue represents the other arranged objects with movable surfaces, and gray represents the environments with fixed surfaces. The label CXFYMZ denotes the number of contacts (X), fixed surfaces (Y), and movable surfaces (Z).
The number of contacts in~\figref{cp} ranges from one (just the floor) to five (floor and all sides). Fixed surfaces refer to those that do not move easily, such as walls and floors, while movable surfaces can move and contact the target object.
Task difficulty is determined by the number and C-patterns, with PP and PT selected accordingly. There are 12 types of C-patterns.

\subsubsection{Defining Policy Patterns}
To determine task policies for PP and PT, we define a policy pattern based on task difficulty and identify the optimal policy by maximizing the sum $f_1+f_2$ of two objectives: accuracy $f_1$ and efficiency $f_2$.

\tabref{pt_rate} shows PT success rates for each C-pattern (number of contacts $N^C$ and fixed surfaces $N^F$). We rank C-patterns by PT success rate in descending order and define each set as a policy pattern. There are 10 policy patterns from $P_0$ (all PP) to $P_9$ (all PT), as illustrated in~\figref{pp}. For the object arrangement operation, we identify the policy pattern that maximizes $f_1+f_2$ among the ten patterns.
\begin{figure}[tb]
    \centering
    \begin{minipage}[tb]{0.50\linewidth}
        \centering
        \includegraphics[keepaspectratio, width=\linewidth]{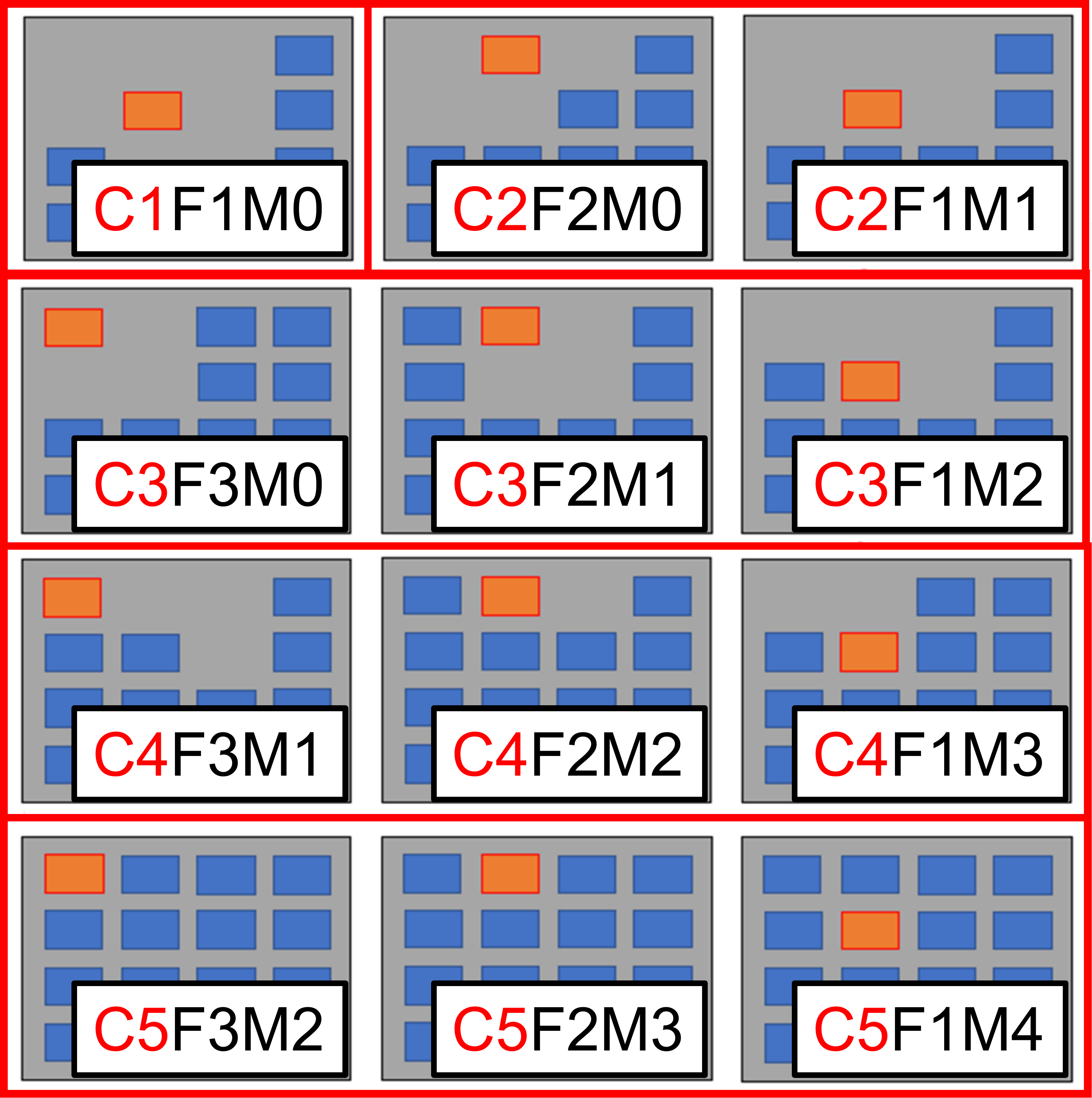}
        \subcaption{\small{Possible C-patterns}}
    \end{minipage}
    \begin{minipage}[tb]{0.46\linewidth}
        \centering
        \includegraphics[keepaspectratio, width=\linewidth]{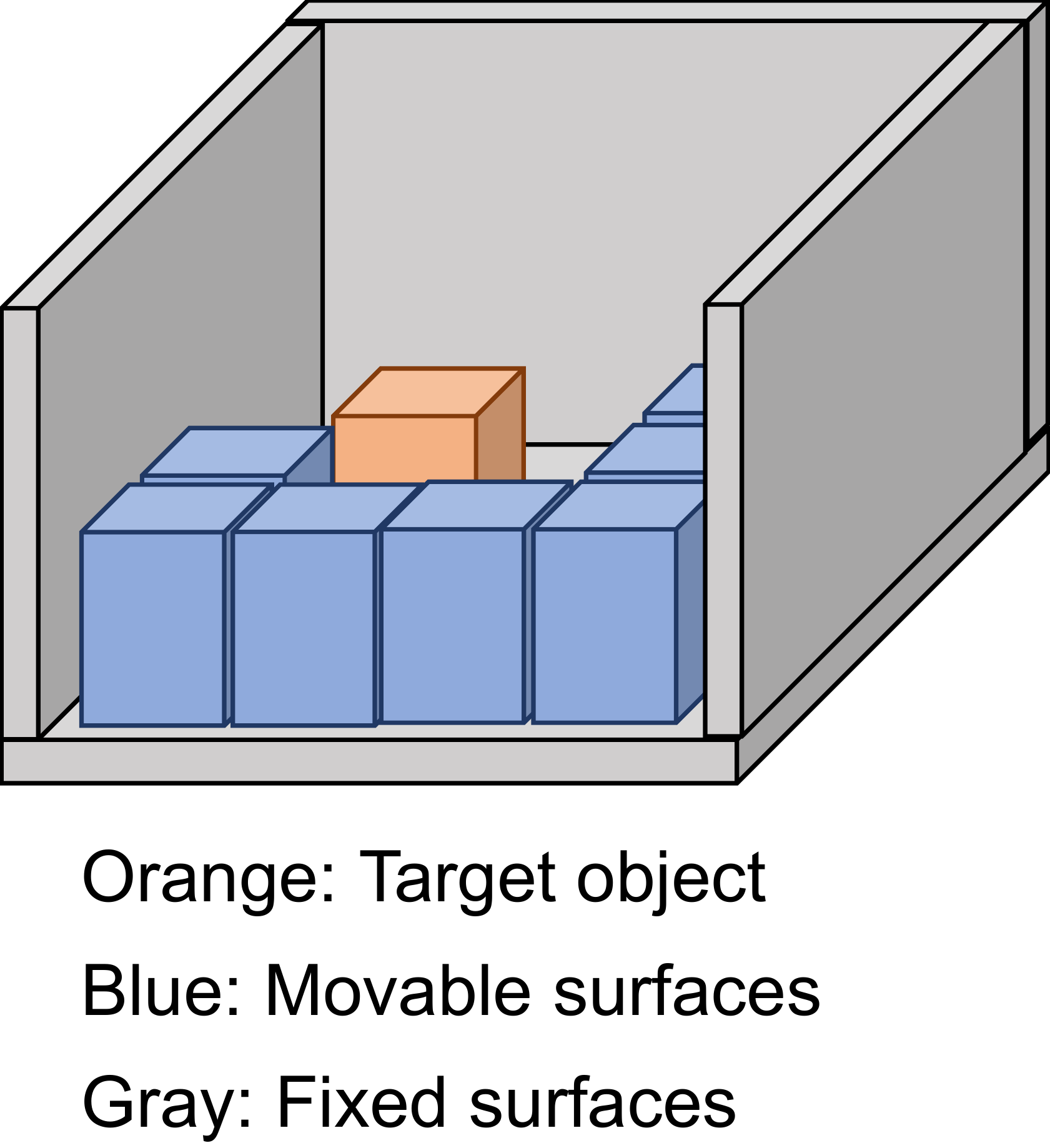}
        \subcaption{\small{Overlooking of C1F1M0}}
    \end{minipage}
    \caption{\small{C-patterns. CXFYMZ represents the numbers of contacts X, fixed surfaces Y, and movable surfaces Z. In the notation CXFYMZ, it always holds that X$=$Y$+$Z.}}
    \figlab{cp}
\end{figure}
\begin{table}[tb]
    \centering
    \small
    \begin{threeparttable}
        \caption{\small{Success rates of PT tasks for different placement environments [\%]}}
        \tablab{pt_rate}
        \begin{tabular}{p{11mm}p{11mm}p{11mm}p{11mm}p{11mm}p{11mm}p{11mm}} \toprule
        \multicolumn{2}{c}{} & \multicolumn{5}{c}{$N^C$ $^{\rm *a}$} \\ \cmidrule(r){3-7}
        \multicolumn{2}{c}{}
            & \multicolumn{1}{c}{1} & \multicolumn{1}{c}{2} & \multicolumn{1}{c}{3} & \multicolumn{1}{c}{4} & \multicolumn{1}{c}{5} \\ \midrule
            & \multicolumn{1}{c}{1} & \multicolumn{1}{r}{~~100~~} & \multicolumn{1}{r}{~~75~~} & \multicolumn{1}{r}{~~50~~} & \multicolumn{1}{r}{~~25~~} & \multicolumn{1}{r}{~~0~~} \\ 
            \multicolumn{1}{c}{$N^F$ $^{\rm *b}$} & \multicolumn{1}{c}{2} & \multicolumn{1}{c}{~~-~~} & \multicolumn{1}{r}{~~100~~} & \multicolumn{1}{r}{~~83~~} & \multicolumn{1}{r}{~~67~~} & \multicolumn{1}{r}{~~43~~} \\ 
            & \multicolumn{1}{c}{3} & \multicolumn{1}{c}{~~-~~} & \multicolumn{1}{c}{~~-~~} & \multicolumn{1}{r}{~~100~~} & \multicolumn{1}{r}{~~75~~} & \multicolumn{1}{r}{~~-~~} \\ \bottomrule
        \end{tabular}
        \begin{tablenotes}
          \item[]\footnotesize{$^{\rm *a}$ Number of contact surfaces}
          \item[]\footnotesize{$^{\rm *b}$ Number of fixed surfaces}
        \end{tablenotes}
    \end{threeparttable}
\end{table}
\begin{figure}[tb]
    \centering
    \includegraphics[keepaspectratio, width=\linewidth]{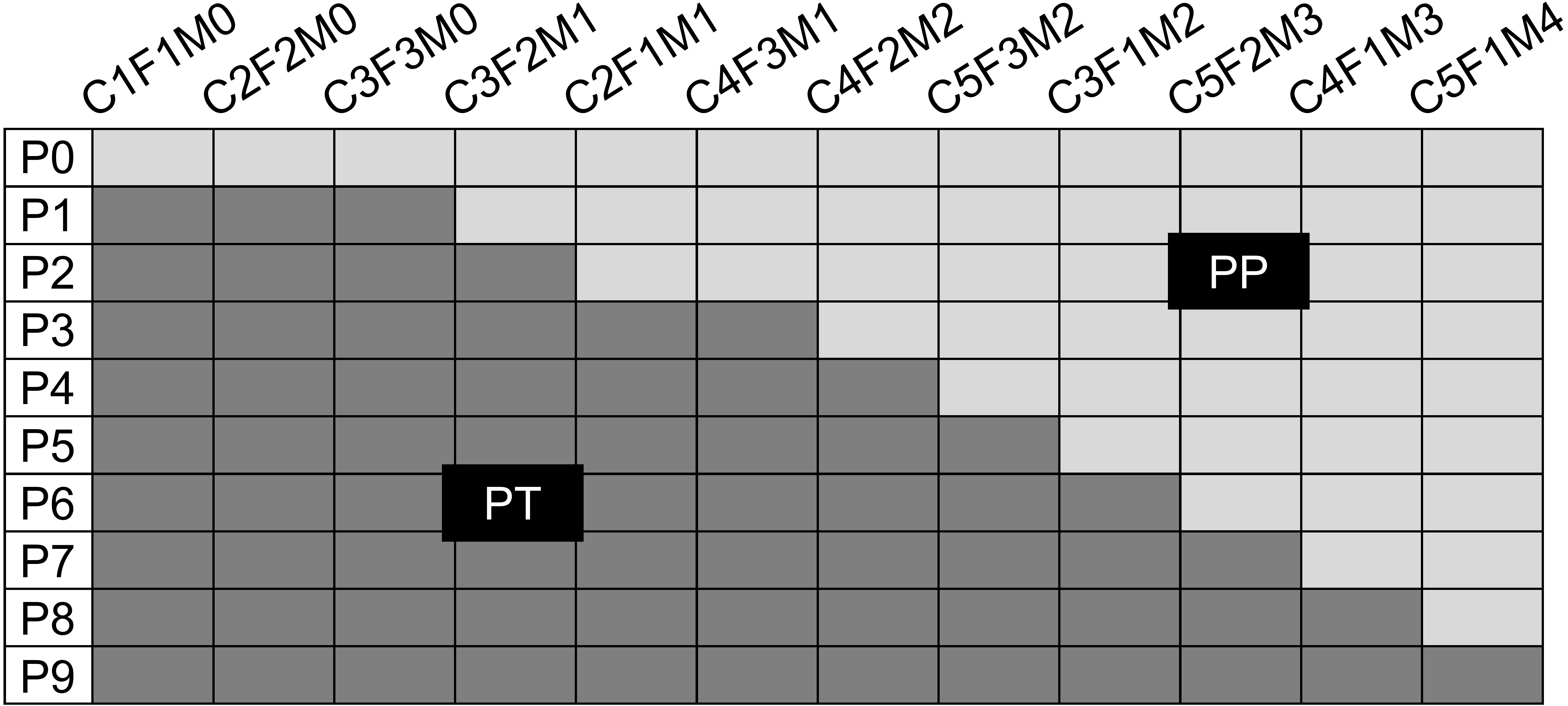}
    \caption{\small{Task assignment table for policy-contact patterns}}
    \figlab{pp}
\end{figure}

\subsubsection{Brute-Force Search for Task Determination Policy}
To determine the PP and PT tasks, a multi-objective optimization problem is solved to find C-patterns and identify the policy pattern that maximizes $f_1+f_2$.

$f_1$ is the normalized success rate of arranging target objects, and $f_2$ is the normalized task completion time. $f_1$ equals 1 if the object is at the target location, and 0 if it exceeds the predetermined distance between adjacent objects from the target. $f_2$ represents the maximum task completion time, with 1 indicating full execution using PP and 0 indicating full execution using PT.

\subsection{Self-supervised Learning of Tossing Motion}
A policy for determining joint angle parameters for a two-fingered gripper executing a tossing motion is derived using self-supervised learning with the $\varepsilon$-greedy method.
This task determination policy is refined through brute-force search. Building on previous self-supervised learning for tossing motions in box packing, we redesign state and reward functions for object arrangement operation.
In the proposed PT, after grasping the object according to its shape, the tossing motion begins at the set arm joint angle, with the object released as the gripper opens to the final width.
Tossing motion learning considers grasping, releasing, and landing posture.

The state $s$ is defined by four dimensions: horizontal grasp position, vertical grasp position, gripper angle at grasping, and gripper opening width at grasping. Action $a$ is defined by the joint angles of the second, third, and fourth joints of the arm with six joints during the tossing motion.
The neural network uses the four-dimensional state space as input, with an action space output of 28$\times$22$\times$16$\times$21$\times$15$\times$18 dimensions, and two 256-dimensional hidden layers. 

Given the wide action space search range, optimal solutions require efficient learning. This is achieved by starting with a tossing motion learned from a reward considering collisions with surrounding objects in the placement area and the appropriate gripper opening width at release, used as prior knowledge.

Action selection during learning uses the $\varepsilon$-greedy method, which chooses the action with the highest Q-value with probability $1-\varepsilon$ and a random action with probability $\varepsilon$, specifically selecting action $\hat{a}$ such that \forref{a}.
\newcommand{\argmax}{\mathop{\rm arg~max}\limits}
\begin{equation} \forlab{a}
    \hat{a}
     =
     \left\{
      \begin{array}{ll}
         \argmax_a Q(s,a), & \text{with probability $1-\varepsilon$} \\
         \text{a random action}, & \text{with probability $\varepsilon$}
        \end{array}
     \right.
\end{equation}
This method facilitates acquiring appropriate Q-values for various actions regardless of the initial Q-value.

We then design the immediate reward value for the tossing motion result, assigning a reward $R_{tr}$ for the success or failure of the release, $R_{lr}$ for the rotation angle of the landing position relative to the target, and $R_{lp}$ for the landing position.
The immediate reward value is the sum of all rewards, as in \forref{R}, and is calculated as follows:
\begin{equation} \forlab{R}
\mathbb{R}=R_{tr}+\alpha R_{lr}+\beta R_{lp},
\end{equation} 
where $\alpha$ and $\beta$ are weighting coefficients. The assignment methods for rewards $R_{tr}$, $R_{lr}$, and $R_{lp}$ are detailed in~\forref{rele}, \forref{deg}, and \forref{posi}.

\forref{rele} provides positive and negative rewards for success or failure, determined empirically.
\forref{deg} assigns a reward of 1 when the object lands at the same angle as the target posture and 0 at the furthest posture, with rewards increasing quadratically as the rotation angle error decreases. 
\forref{posi} assigns a reward of 1 when the object lands within a non-collision range and 0 when it lands more than the width of a target object away, with rewards increasing linearly as the distance to the target position decreases.
\begin{equation} \forlab{rele}
    R_{tr}=
        \left\{
            \begin{array}{ll}
                1 & \text{(release success)} \\
                -10 & \text{(release failure)}
            \end{array},
        \right.
\end{equation}
\begin{flalign}
    \begin{aligned} \forlab{deg}
        R_{lr}&=r_{roll}+r_{yaw},& \\
        r_{roll}&=
        \left\{
            \begin{array}{ll}
                1/\bar{\theta}_{roll}^2 (\theta_{roll}-\bar{\theta}_{roll})^2 & \text{($\theta_{roll}\leq\bar{\theta}_{roll}$)} \\
                0 & \text{($\theta_{roll}>\bar{\theta}_{roll})$} \\
            \end{array},
        \right.&\\
        r_{yaw}&=
        \left\{
            \begin{array}{ll}
                1/\bar{\theta}_{yaw}^2 (\theta_{yaw}-\bar{\theta}_{yaw})^2 & \text{($\theta_{yaw}\leq\bar{\theta}_{yaw}$)} \\
                0 & \text{($\theta_{yaw}>\bar{\theta}_{yaw}$)} \\
            \end{array}.
        \right.&
    \end{aligned}
\end{flalign}
\begin{flalign}
    \begin{aligned} \forlab{posi}
        R_{lp}&=r_{x}+r_{y}+r_{z},& \\
        r_{x}&=
        \left\{
            \begin{array}{ll}
                1 & (d_{x} \leq d_{x}^{l}) \\
                -1/(d_{x}^{h} - d_{x}^{l})d_{x} + d_{x}^{h}/(d_{x}^{h} - d_{x}^{l}) & (d_{x}^{l} < d_{x} \leq d_{x}^{h}) \\
                0 & (d_{x} > d_{x}^{h})
            \end{array},
        \right.&\\
        r_{y}&=
        \left\{
            \begin{array}{ll}
                1 & (d_{y} \leq d_{y}^{l}) \\
                -1/(d_{y}^{h} - d_{y}^{l})d_{y} + d_{y}^{h}/(d_{y}^{h} - d_{y}^{l}) & (d_{y}^{l} < d_{y} \leq d_{y}^{h}) \\
                0 & (d_{y} > d_{y}^{h})
            \end{array},
        \right.&\\
        r_{z}&=
        \left\{
            \begin{array}{ll}
                0 & (d_{z} \geq \bar{d}_{z}) \\
                -1 & (d_{z} < \bar{d}_{z}) \\
            \end{array}.
        \right.&
    \end{aligned}
\end{flalign}
In this context, \text{release success} occurs when the target object is released from the gripper and tossed forward, while \text{release failure} happens if the object falls to the ground while still held by the gripper or before the arm reaches the release position.

$\theta_{roll}$ and $\theta_{yaw}$ are the roll and yaw angles from release to landing, and $d_{x}$, $d_{y}$, and $d_{z}$ indicate the distances between the landing and target positions in the $x$, $y$, and $z$ directions.
These directions correspond to the $XYZ$ axes shown in~\figref{overview}. $\bar{\theta}_{roll}$ and $\bar{\theta}_{yaw}$ are thresholds for rewarding errors in each rotation angle. $d_{x}^{l}$ and $d_{y}^{l}$ ensure a high reward area where the robot can land without hitting surrounding objects. $d_{x}^{h}$ and $d_{y}^{h}$ are based on the distance to the target object, and $\bar{d}_{z}$ is set to the height of the target placement position.

\section{Training Results in Simulations}
\begin{figure}[tb]
    \centering
    \includegraphics[keepaspectratio, width=\linewidth]{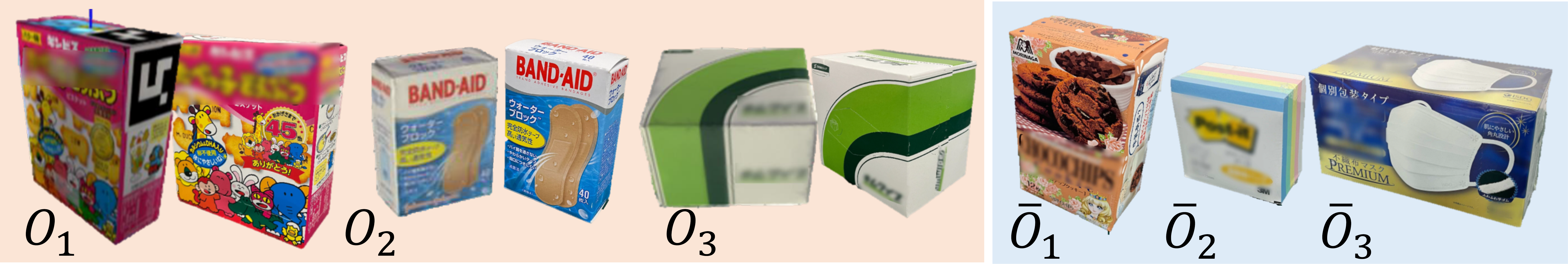}
    \vspace{0.2mm}
    \caption{\small{Target objects. $O_i~(i\in\{1,2,3\})$ and $\bar{O}_j~(j\in\{1,2,3\})$ are the objects used (Trained) and not used (Unknown) in training}}
    \figlab{object}
  \end{figure}
\begin{figure}[tb]
    \centering
    \begin{minipage}[tb]{\linewidth}
        \centering
        \includegraphics[keepaspectratio, width=0.9\linewidth]{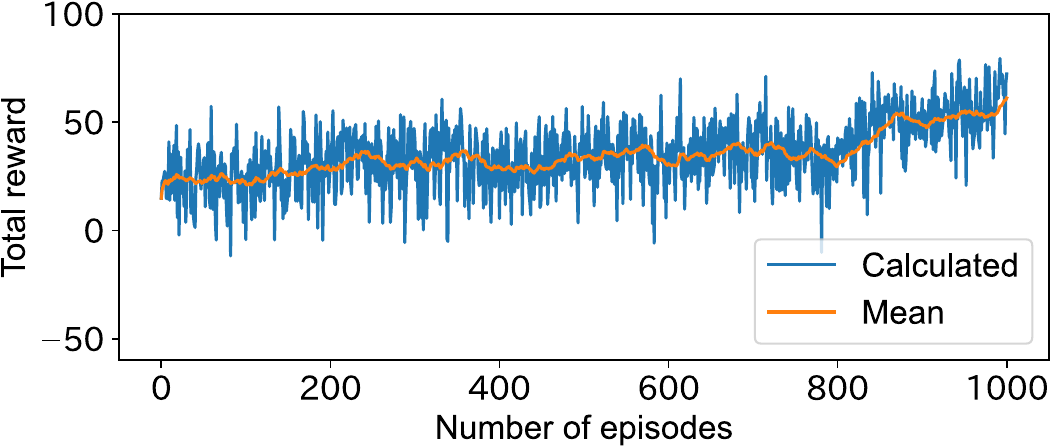}
        \subcaption{\small{Total reward values through 1000 learning episodes for $O_1$}}
    \end{minipage}
    \begin{minipage}[tb]{\linewidth}
        \centering
        \includegraphics[keepaspectratio, width=0.8\linewidth]{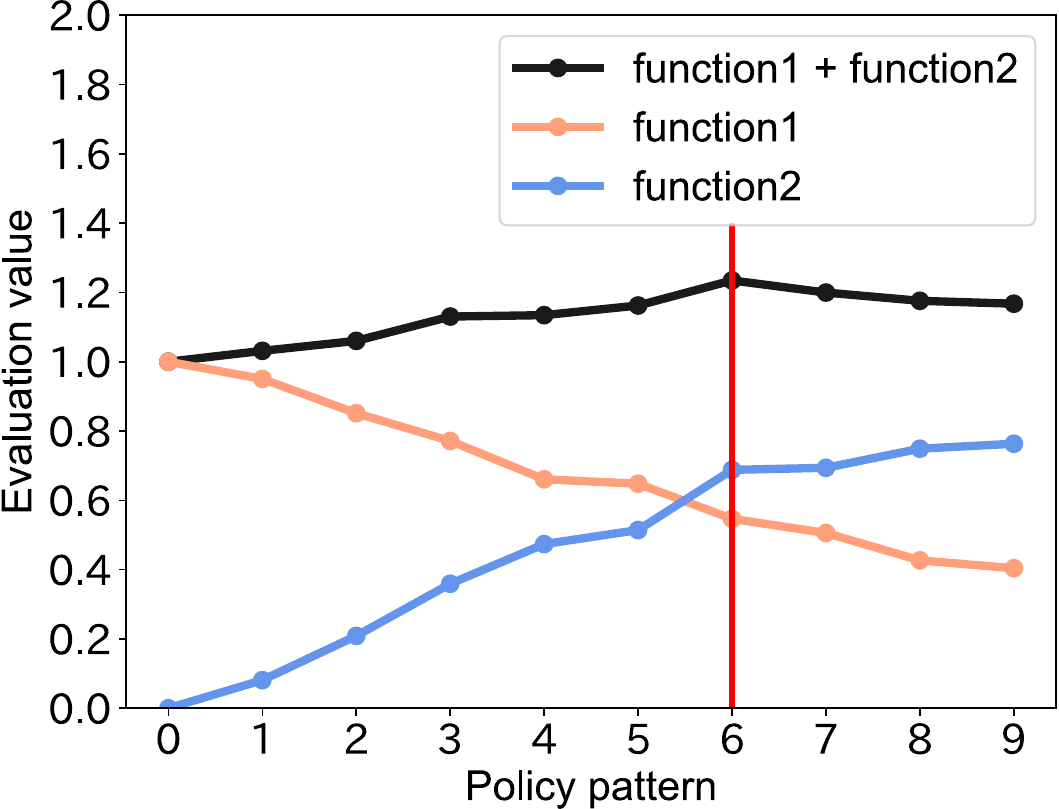}
        \subcaption{\small{Evaluation values for the policy patterns for $O_1$}}
    \end{minipage}
    \caption{\small{Training results. In (a), the blue lines show the calculated values of the rewards and the orange lines show the mean values. In (b), the red line shows the determined policy pattern.}}
    \figlab{training}
\end{figure}

As illustrated in~\figref{overview}, the experimental setup involved arranging placements and containers using a robotic arm (UR5e, Universal Robots) with a two-finger gripper (2F-140, Robotiq).
Three types of rectangular shapes $O_i~(i\in\{1,2,3\})$ with different sizes were used as target objects, as shown in~\figref{object}. The base width$\times$depth$\times$height [mm] of each $O_i$ were 116$\times$49$\times$135, 65$\times$30$\times$93, and 133$\times$88$\times$120,~respectively.

For DQN training, we used the following parameters: 1000 episodes, 30 steps per episode, initial $\varepsilon$ of 0.95 decreasing to 0.05, a discount rate of 0.8, a learning rate of 0.001, and a batch size of 64. The brute-force search for task determination policies for PP and PT is computationally prohibitive. For example, arranging 20 target objects in a $4\times5$ matrix yields $2^{21}\times5$ possible environments. Thus, we constrained the task to 48 environments, excluding similar configurations, to include all possible C-patterns.

We computed the mean value of $f_1+f_2$ for the object arrangement task for each policy pattern in~\figref{pp} across these environments. Parameters $\alpha$ and $\beta$ in \forref{R} were set to one. Values for $\bar{\theta}_{roll}$, $\bar{\theta}_{yaw}$, $d_{x}^{l}$, $d_{y}^{l}$, $d_{x}^{h}$, $d_{y}^{h}$ and $\bar{d}_{z}$ in \forref{rele}, \forref{deg}, and \forref{posi} were 360, 180, 0.03, 0.03, 0.15, 0.06, and 0.3, respectively. 

\figref{training}~(a) illustrates the immediate reward value transition during learning for $O_1$, with similar trends observed for $O_2$ and $O_3$. The reward value increases with learning steps, validating the learning process.

\figref{training}~(b) shows the task determination policy results for $O_1$, with similar trends for $O_2$ and $O_3$. Function1 represents accuracy $f_1$, and function2 denotes efficiency $f_2$; their sum, $f_1 + f_2$, is highlighted in red at the maximum value. The general-purpose policy pattern for all target objects was determined as $P_6$, balancing accuracy and efficiency, which are trade-offs.

\section{Real-World Feasibility}
To check the feasibility in the real world, using the final estimated values of the joint angle parameters at the time of release obtained after learning as a reference, 10 trials were conducted for each of the three different C-patterns (C4F4M2, C5F3M2, and C3F1M2) for a rectangular shape in a real environment.
In addition, we confirmed the effectiveness of the task determination policy with $P_6$ by comparing the task success rate and task completion time with those of the other policies with $P_3$ and $P_9$.

The objects used in the experiment included three types of objects $O_i$ in~\figref{object} for training (Trained) and three types of objects $\bar{O}_j$ for simulation (Unknown), totaling six types. The dimensions of the unknown objects were 94$\times$59$\times$157, 75$\times$43$\times$75, and 200$\times$100$\times$103 (width$\times$depth$\times$height [mm]).
For the tossing motion parameters of the unknown object, parameters for a trained object of similar shape (same category and smallest squared error for three sides) were used based on Chen~\etal~\cite{Chen2022}.

\begin{figure}[tb]
    \centering
    \begin{minipage}[tb]{\linewidth}
        \centering
        \includegraphics[keepaspectratio, width=\linewidth]{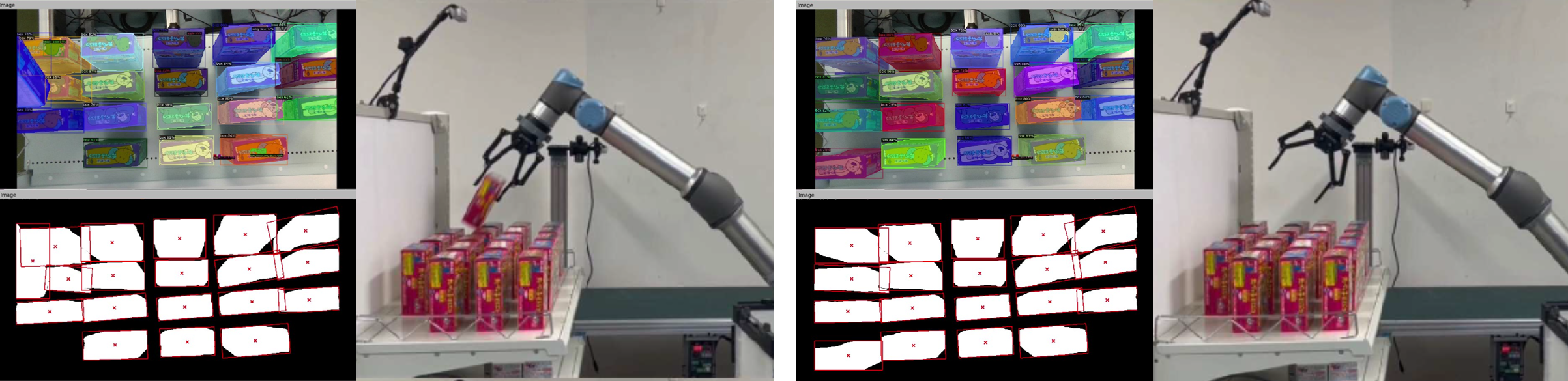}
        \subcaption{\small{$O_1$ in C5F3M2}}
    \end{minipage}
    \begin{minipage}[tb]{\linewidth}
        \centering
        \includegraphics[keepaspectratio, width=\linewidth]{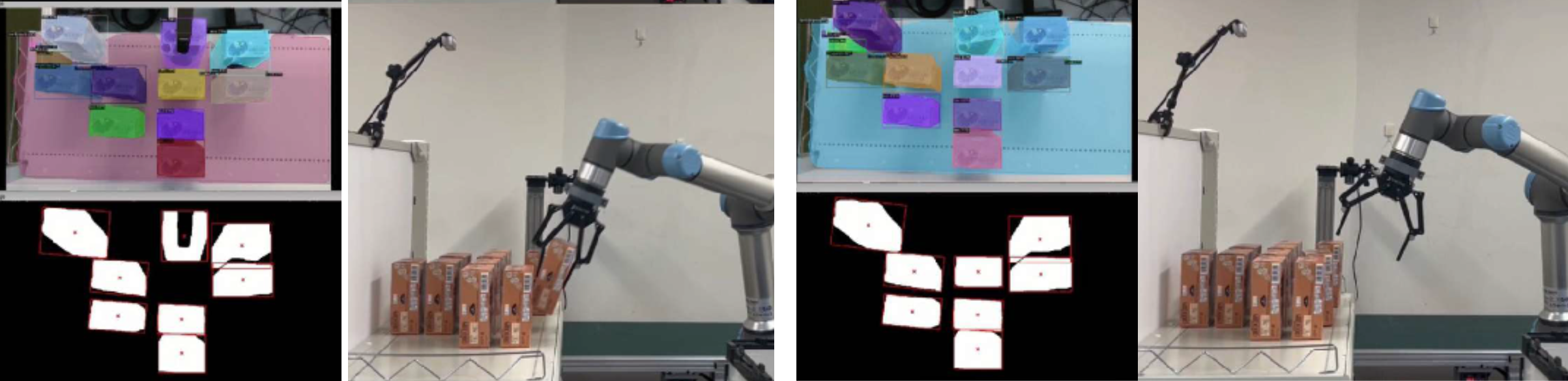}
        \subcaption{\small{$\bar{O}_1$ in C4F1M3}}
    \end{minipage}
    \vspace{1mm}
    \caption{\small{Real-world tossing motions for the trained object in C5F3M2 (a)~$O_1$ and unknown object (b)~$\bar{O}_1$ in C4F1M3.}}
    \figlab{real}
\end{figure}
\begin{table}[tb]
    \centering
    \small
    \begin{threeparttable}
        \caption{\small{Success rates of real-world tossing motions [\%]}}
        \tablab{result_real}
        \begin{tabular}{p{9mm}p{9mm}p{9mm}p{9mm}p{9mm}p{9mm}p{9mm}} \toprule
            & \multicolumn{3}{c}{Trained} & \multicolumn{3}{c}{Unknown} \\
            \cmidrule(r){2-7}
            \multicolumn{1}{c}{C-pattern} & \multicolumn{1}{c}{$O_1$} & \multicolumn{1}{c}{$O_2$} & \multicolumn{1}{c}{$O_3$} & \multicolumn{1}{c}{$\bar{O}_1$} & \multicolumn{1}{c}{$\bar{O}_2$} & \multicolumn{1}{c}{$\bar{O}_3$} \\ \midrule
            \multicolumn{1}{c}{C4F2M2} & \multicolumn{1}{r}{~100~~} & \multicolumn{1}{r}{~70~~} & \multicolumn{1}{r}{~100~~} & \multicolumn{1}{r}{~90~~} & \multicolumn{1}{r}{~90~~} & \multicolumn{1}{r}{~100~~} \\ 
            \multicolumn{1}{c}{C5F3M2} & \multicolumn{1}{r}{~100~~} & \multicolumn{1}{r}{~60~~} & \multicolumn{1}{r}{~90~~} & \multicolumn{1}{r}{~90~~} & \multicolumn{1}{r}{~70~~} & \multicolumn{1}{r}{~80~~} \\
            \multicolumn{1}{c}{C3F1M2} & \multicolumn{1}{r}{~100~~} & \multicolumn{1}{r}{~70~~} & \multicolumn{1}{r}{~90~~} & \multicolumn{1}{r}{~80~~} & \multicolumn{1}{r}{~80~~} & \multicolumn{1}{r}{~80~~} \\ \bottomrule
        \end{tabular}
    \end{threeparttable}
\end{table}

\figref{real}~(a) shows the tossing motion using a trained object $O_1$ in C5F3M2, while \figref{real}~(b) displays the motion with an unknown object $\bar{O}_1$ in C4F1M3. The robotic arm successfully released the object from the specified position and posture at the release joint angle output from the learned model, confirming that the object arrangement task is feasible using a tossing motion similar to projectile motion.

\tabref{result_real} shows the success rate of the tossing motion in 10 trials for each target object and environment. The success rate was higher for $O_1$ and $O_3$ compared to any unknown object, suggesting effective learning. For $O_2$, which was smaller and thinner, the tossing motion was more challenging.
Although the success rates for $\bar{O}_1$, $\bar{O}_2$, and $\bar{O}_3$ are lower than for trained objects $O_1$ and $O_3$, they still maintain a minimum of 70\%.
The average success rate for the three types of unknown objects in three environments was high at 84\%, demonstrating applicability to unknown objects by referencing similar shapes.

In addition, 10 trials were performed on the object placement task using PP and PT with the task determination policies of P3, P6, and P9.
12 objects were randomly placed in the workspace for the initial placement, and the task was to place the remaining 8 objects.
As a result, the average success rate and task completion time for the 8 tasks of P3, P6, and P9 were 112 seconds for 88\%, 80 seconds for 86\%, and 76 seconds for 68\%, respectively.
The time required to repair failed tasks was not included in the task completion time.
The results were comparable to those of the simulation, suggesting that the task determination policy of P6 can achieve balanced tasks in terms of success rate and task completion time in the real world.

\section{Conclusions}
Efficient robot operation in object arrangement is essential. This study introduces pick-and-toss (PT) to replace pick-and-place (PP), extending the robot's range and task efficiency. PT enhances efficiency, but environmental factors affect tossing success. Cluttered areas with many movable surfaces require caution to avoid collisions, whereas fixed surfaces like walls simplify the task. This study aims for accurate and efficient object arrangement by determining PP and PT tasks, considering placement environment difficulty. We proposed a method to simultaneously learn tossing motion through self-supervised learning and establish a task determination policy via brute-force search, proving effective in simulations and real-world tests.

Future work will focus on recognizing the grasping state affecting tossing motion and developing a self-supervised learning approach from limited real-world samples. With only 12 C-patterns currently available, we plan to create an open-ended model for PP and PT policy determination for additional C-patterns.

\bibliographystyle{IEEEtran}
\footnotesize
\bibliography{reference}

\end{document}